\documentclass[11pt,a4paper]{article}
\pdfoutput=1

\usepackage[english]{babel}
\usepackage[utf8x]{inputenc}
\usepackage[T1]{fontenc}

\usepackage{graphicx}
\graphicspath{{../Figures/}{./}}

\usepackage{enumerate}
\usepackage{enumitem}
\usepackage{url} 
\usepackage[ruled,lined]{algorithm2e}

\usepackage[margin=1in]{geometry}

\let\bibhang\relax
\usepackage[authoryear]{natbib}
\usepackage{booktabs}
\usepackage{amsmath, amsthm, amssymb, bm}

\usepackage{gensymb}
\usepackage{subcaption}
\usepackage{multicol}
\DeclareMathOperator{\E}{\mathbb{E}}
\DeclareMathOperator{\diag}{diag}
\newcommand{\dif}[1]{\ensuremath{\operatorname{d}\!{#1}}}
\newcommand{\T}{^{\top}}
\newcommand{\X}{\boldsymbol{X}}
\newcommand{\x}{\boldsymbol{x}}
\newcommand{\Z}{\boldsymbol{Z}}
\newcommand{\z}{\boldsymbol{z}}
\newcommand{\mub}{\boldsymbol{\mu}}
\newcommand{\Sigmab}{\boldsymbol{\Sigma}}
\newcommand{\Psib}{\boldsymbol{\Psi}}
\newcommand{\A}{\boldsymbol{A}}
\newcommand{\D}{\boldsymbol{D}}
\renewcommand{\S}{\boldsymbol{S}}

\newcommand{\B}{\boldsymbol{B}}
\renewcommand{\b}{\boldsymbol{b}}
\renewcommand{\t}{\boldsymbol{t}}

\newcommand{\I}{\boldsymbol{I}}
\newcommand{\1}{\boldsymbol{1}}
\renewcommand{\hat}[1]{\widehat{#1}}
\newcommand{\MC}{\mathrm{MC}}
\newcommand{\Rel}{\mathrm{Rel}}
\newcommand{\UT}{\mathrm{UT}}
\newcommand{\VAR}{\mathrm{VAR}}
\newcommand{\SOTE}{\mathrm{SOTE}}
\newcommand{\SNR}{\mathrm{SNR}}
\newcommand{\Signal}{\mathrm{S}}

\makeatletter
\newcommand{\verbatimfont}[1]{\def\verbatim@font{#1}}%
\makeatother

\makeatletter
\newcommand\code{\bgroup\@makeother\_\@makeother\~\@makeother\$\@codex}
\def\@codex#1{{\normalfont\ttfamily\hyphenchar\font=-1 #1}\egroup}
\makeatother
\newcommand{\pkg}[1]{{\fontseries{b}\selectfont #1}}

\begin{document}


\title{Projection pursuit based on Gaussian mixtures 
       and evolutionary algorithms}
\author{%
  Luca Scrucca \\
  Department of Economics, University of Perugia\\
  and \\
  Alessio Serafini \\
  Department of Economics, University of Perugia}
\date{\today}
\maketitle

\begin{abstract}
We propose a projection pursuit (PP) algorithm based on Gaussian mixture models (GMMs).
The negentropy obtained from a multivariate density estimated by GMMs is adopted as the PP index to be maximised. For a fixed dimension of the projection subspace, the GMM-based density estimation is projected onto that subspace, where an approximation of the negentropy for Gaussian mixtures is computed. Then, Genetic Algorithms (GAs) are used to find the optimal, orthogonal projection basis by maximising the former approximation.
We show that this semi-parametric approach to PP is flexible and allows highly informative structures to be detected, by projecting multivariate datasets onto a subspace, where the data can be feasibly visualised. The performance of the proposed approach is shown on both artificial and real datasets. \\

\noindent{\it Keywords:} data visualisation, dimension reduction method, clustering visualisation, multivariate data analysis, clustering analysis.
\end{abstract}

\newpage
\baselineskip=18pt

\section{Introduction}

Exploratory data analysis is an important step to understand the underlying structure of the data.
Where embedded phenomena cannot be directly visualised for multivariate datasets, graphical inspection techniques represent a useful tool.
Thus, dimension reduction methods were developed to project the data in a feasible space, where they could be visualised. Principal component analysis and factor analysis are classical dimension reduction procedures. The first takes into account the variability contained in the features and it is widely used as a result of its straightforward computability and well understood objectives, whereas the second one uses the correlation between the features.

Following the idea of \citet{Kruskal:1969}, \citet{Friedman:Tukey:1974} implemented a features extraction method they named projection pursuit (PP). This is a dimension reduction technique which investigates high-dimensional data to find ``interesting'' low-dimensional linear projections. It does so by maximising an index, known as projection index, to discover the best orthogonal projections, which incorporate the information on the ``interesting'' directions.
The definition of ``interesting'' may differ, depending on the applications or purposes: for example, the PP method has been applied to regression, density estimation and supervised learning \citep{Huber:1985}.

The purpose of this work is to present a PP method capable of visualising clustering structures. Since Gaussian projections are likely to be the less informative projections for clustering visualisation purposes, departures from the normality are considered as a definition of ``interesting'' projections.

However, most low-dimensional projections are approximately normal \citep{Diaconis:Freedman:1984, Huber:1985}. 
Several different PP indices, which seek departure from normality, have been proposed, such as the entropy index and a moment index \citep{Jones:Sibson:1987}, indices based on the $L^2$ distance, e.g. Legrende index \citep{Friedman:1987}, Hermite index \citep{Hall:1989} and Natural Hermite index \citep{Cook:Buja:Cabrera:1993}, and distance-based indices, e.g. based on the Chi-square distance \citep{Posse:1995}.

A general PP procedure can be summarised as follows:
\begin{enumerate}
  \item Transform the data (usually sphering).
  \item Choose the PP index.
  \item Set the dimension of the subspace where the data must be projected.
  \item Optimise the PP index using some maximisation algorithms.
\end{enumerate}
In the PP framework, the dimension of the projected subspace is usually fixed and it depends on the purpose of the application. If a visual inspection is performed, then one and two dimensions are the most common choices.

We propose a new PP method, which employs Genetic Algorithms (GAs) to maximise negentropy, which is a measure of distance to normality. In our case, negentropy is derived with respect to the Gaussian mixture distribution of the projected data. We refer to this approach as PPGMMGA.

GMMs are a powerful tool for semi-parametric density estimation and many densities can be approximated by GMMs, especially if the data tends to have a multimodal structure. In addition to providing an approximation of complex distributions, GMMs can also be used for clustering purposes. If the underlying structure of the data contains any cluster structure, GMMs could be able to detect it, providing a classification of observations and estimation of the parameters. Furthermore, when the density is estimated by means of GMMs on the original data, the projected data density is easily calculated thanks to the linear transformation property of Gaussian mixture models.

Most of the PP indices of non-normality are based on non-parametric density estimation of the projected data. If GMMs are used to estimate density, then the linear transformation property immediately provides the density of the projected data. Optimising this PP index can be very difficult due to the possible multimodal nature of the objective function surface. Thus, many optimisation methods are unsuitable in the case of multiple optima, and stochastic algorithms, such as GAs, may be a viable alternative for the non-convex nature of the objective function.

Since the negentropy index \citep{Huber:1985, Jones:Sibson:1987} is based on the entropy of the projected data density, and since the entropy of GMMs does not have a closed formula, the index is approximated, and then optimised via GAs, to discover the best basis that describes the least normality. Different approximations are presented and discussed.

The steps of the proposed procedure are summarised in Algorithm~\ref{algo:PPGMMGA}.

\SetKwInOut{Steps}{Steps}
\begin{algorithm}[ht]
\linespread{1.1}\selectfont
\caption{PPGMMGA}
\label{algo:PPGMMGA}
\KwIn{%
\begin{enumerate}[nosep, label={--}]
\item Data matrix $\X$ containing the values for $n$ observations/units on $p$ variables/features.
\item The dimension $d$ of the projection subspace.
\end{enumerate}}
\Steps{%
\begin{enumerate}[nosep]
\item Estimate the density of the centred, and possibly scaled, data using GMMs (Section~\ref{sec:densityPP}).
\item By exploiting the linear transformation property of GMMs (Section~\ref{sec:gmmlinproj}), maximise an approximated negentropy index (Section~\ref{sec:negentprojind}) using GAs (Section~\ref{sec:maxgmmnegent}) as optimisation tool.
\end{enumerate}}
\KwOut{%
\begin{enumerate}[nosep, label={--}]
\item Matrix $\hat{\B}$ of size $p\times d$, providing the estimated basis of the projection subspace.
\item Data matrix $\hat{\Z}$ of size $n\times d$, representing the projection of points onto the estimated subspace (Section~\ref{sec:gmmlinproj}).
\end{enumerate}}
\end{algorithm}

The paper is organised as follows. In Section~\ref{sec:background} we introduce the necessary background. Then, in Section~\ref{sec:methodology}, we discuss the proposed projection pursuit procedure, by demonstrating how to calculate a PP index from a density estimated with Gaussian mixtures, using closed-form approximations to compute the negentropy, and maximising the chosen PP index with GAs. Section~\ref{sec:examples} presents some applications of the proposed methodology on real and simulated data. The final section provides some concluding remarks.

\section{Background}
\label{sec:background}

\subsection{Density estimation by GMMs}
\label{sec:densityPP}

Let $\x$ be a random vector in $\mathbb{R}^p$ with density $f(\x)$, which can be described by a mixture of Gaussian distributions. The last assumption implies that the density can be written as follows:
\begin{equation}
f(\x ; \Psib) = \sum^{G}_{g=1} \pi_{g} \phi( \x ; \mub_g, \Sigmab_g),
\label{eq:gmm}
\end{equation}
where $\Psib = \{\pi_1,\pi_2, \dots , \pi_{G-1}, \mub_1, \dots ,  \mub_G,  \Sigmab_1, \dots, \Sigmab_G \}$ are the parameters of the mixture model, with ($\pi_1,\pi_2, \dots, \pi_G$) the mixing weights, so that $\pi_g > 0$ and $\sum_{g=1}^{G} \pi_g = 1$, $G$ is the number of components, and $\phi(\x ;  \mub_g, \Sigmab_g)$ the underlying multivariate density function of \textit{g}th component with parameters $ \mub_g$, $\Sigmab_g$.
Parsimonious parametrisations of covariance matrices for GMMs can be obtained using the following eigen-decomposition $\Sigmab_g = \lambda_g\D_g\A_g\D\T_g$ \citep{Banfield:Raftery:1993, Celeux:Govaert:1995}, where $\lambda_g$  is a scalar controlling the volume of the ellipsoid, $\A_g$ is a diagonal matrix controlling its shape, and $\D_g$ is an orthogonal matrix controlling the orientation of the ellipsoid.
GMMs can approximate any continuous density with arbitrary accuracy provided the model has a sufficient number of components and the parameters of the model are correctly estimated \citep{Escobar:West:1995, Roeder:Wasserman:1997}.

A standard algorithm to obtain maximum likelihood estimates of mixture parameters is the Expectation-Maximization (EM) algorithm \citep{Dempster:Laird:Rubin:1977, McLachlan:Krishnan:2008}. The EM is an iterative algorithm, which generates a sequence of parameters estimates by alternating two steps. The expected value of complete-data log-likelihood is computed (E-step) and the parameters are updated by maximising (M-step) the expectation computed in the previous step. Then, it can be shown that the EM algorithm converges at least to a local optimum under fairly general conditions \citep{McLachlan:Krishnan:2008}.

Information criteria based on penalised forms of the log-likelihood are routinely used in GMMs for model selection, i.e. to decide both how many components should be included in the mixture and which covariances parameterisation to adopt. Standard criteria for model selection are the Bayesian information criterion \citep[BIC; ][]{Schwarz:1978} and the integrated complete-data likelihood \citep[ICL; ][]{Biernacki:Celeux:Govaert:2000}.

\subsection{Centring and scaling of the input data}
\label{sec:center}

Let $\x_i$ be the $i$th observation drawn from the distribution of $\x$, and collect the $n$ random sample points in the $n \times p$ matrix $\X$.
As suggested by \citet{Jones:Sibson:1987}, before applying a PP algorithm it is customary to sphere the data, i.e. computing the transformation $\X \gets (\X - \1_n\mub\T)\Sigmab^{-\frac{1}{2}}$, where $\1_n$ is the unit vector of length $n$, $\mub$  and $\Sigmab$ are respectively the mean vector and the covariance matrix of $\X$. Such transformation makes the new sphered data centred to zero, with unit variance, and zero covariances. However, as noted by both \citet{Gower:1987}, and \citet{Hastie:Tibshirani:1987}, the sphering process changes the shape of the data, and it may hide interesting structures, otherwise visible without sphering.

Semi-parametric density estimation via GMMs can be influenced by data sphering. Constraining the covariance matrix to be unitary may reduce the information available, hence degrading the density estimation fit. In some cases, the estimation of GMMs when the data are sphered cannot detect any clustering structure.

Thus, in our approach we pre-process the data differently. A preliminary centring step is always performed, so that the processed data are centred at zero, i.e.
\begin{equation*}
\X \gets (\X - \1_n\mub\T),
\end{equation*}
where $\mub$ is the vector of means for each variable. Thus, the origin of the axes in the projection space coincides with the centroid of the data points.

Scaling is a further step that is advisable if the scale of the variables is supposed to influence the analysis, because the variables are expressed in different units of measure or take values in very different ranges. Centring and scaling of the data is obtained by the following transformation:
\begin{equation*}
\X \gets (\X-\1_n\mub\T)\S^{-1/2},
\end{equation*}
where $\S = \diag(\sigma^2_1, \ldots, \sigma^2_p)$ is the diagonal matrix containing the variances of each variable. 
Note, however, that such scaling preserves the correlation structure among the variables.

\subsection{Entropy}
\label{sec:entropy}

Entropy for continuous random variables, also called differential entropy, is an extension of the classic entropy introduced by \citet{Shannon:1948}. This is a measure of uncertainty or information content in a random variable. For a random vector $\x \in \mathbb{R}^p$ with probability density function $f(\x)$, the entropy is defined as follows:
\begin{equation}
\label{E}
h(\x) = - \E_f[ \log f(\x)] = - \int \log\big(f(\x)\big) f(\x) \dif\x.
\end{equation}
A closed formula for entropy is available if $\x \sim N(\mub, \Sigmab)$ \citep{Cover:Thomas:2006}, namely
\begin{equation*}
\label{E:G}
h(\x) = \frac{1}{2} \log \big( (2\pi e)^d |\Sigmab|\big).
\end{equation*}

On the contrary, a closed formula for entropy does not exist for GMMs, thus an approximation is required.
A technique that directly approximates  equation \eqref{E} is the Monte Carlo (MC) method, where we draw  i.i.d samples $\x_i$ $(i = 1,\dots,S)$  from $f(\x)$ and compute:
\begin{equation*}
h_{\MC}(\x) = -\frac{1}{S} \sum_{i = 1}^S \log f(\x_i).
\end{equation*}
It is straightforward to note that, by the law of large numbers, $h_{\MC}(\x) \to h(\x)$ as $S \to \infty$.
MC approximation guarantees convergence to the true value of the entropy, but a very large MC sample size is required to reasonably approximate the expected value.

\section{Methodology}
\label{sec:methodology}

\subsection{Distribution of linear projections of GMMs}
\label{sec:gmmlinproj}

Capturing cluster structure in a dataset is not the only useful GMMs characteristic for PP purposes. One useful feature exploited in this work is the \textit{linear transformation property of GMMs}. If the data is projected using linear transformation and a GMM is fitted on the original data, then the density on the projection subspace is again a mixture of Gaussian variables with covariance matrices and mean vectors obtained by a linear transformation of the parameters estimated in the original scale of the data.

Consider an orthogonal linear projection $L:\mathbb{R}^{p} \rightarrow \mathbb{R}^{d}$, with $p \ge d$, i.e. any linear mapping such that:
\begin{equation*}
\z = \B\T \x  \in \mathbb{R}^{d},
\end{equation*}
with $\B \in \mathbb{R}^{p \times d}$, where the column vectors of $\B$ are orthogonal to each other and with unit length, and $\x$ distributed according to the Gaussian mixture model in \eqref{eq:gmm}. As shown in the Appendix, $\z$ is again a Gaussian mixture with density:
\begin{equation}
\label{mixtureLinear}
f(\z) = \sum^{G}_{g=1} \pi_{g}\phi( \z ; \B\T \mub_g, \B\T \Sigmab_g \B).
\end{equation}
This property allows the density of the original data to be easily projected onto a lower dimensional subspace, by transforming only the component means and the component covariance matrices with the basis matrix $\B$, hence making the GMMs tractable for PP purposes.

Finally, the new data cloud $\Z = \X\B$ represents the $n$ sample points projected in the lower dimensional subspace spanned by the basis matrix $\B$.

\subsection{Negentropy as a projection index}
\label{sec:negentprojind}

The choice of projection index is the most important step in any PP procedure. This index must incorporate the information on how ``interesting'' a projection is and translate it into a numerical value.
\citet{Huber:1985} proposed a theoretical framework for projection indices, classifying them in three different categories. In a PP framework, the projection indices should be affine invariant, as they are not affected by changing the data location and scaling. Thus, the subspace generated is unique, and the solution space in the optimisation problem is the same for all affine transformations.

If we take the adopted definition of ``interesting'' projections to be those showing non-normality, and recall any affine transformation of Gaussian is again Gaussian, the affine invariant property guarantees that the objective function is constant for all the Gaussian distributions with the same dimensionality. Thus, the PP index is the same for each basis, which generates Gaussian projections of the data.

In addition to this property, the index should be efficient from a computational point of view, and tractable from a theoretical point of view.

One choice for the PP index might be to directly employ the entropy on the density of the projected data. Nevertheless, such index is not invariant to scale transformations \citep{Cover:Thomas:2006}, which makes it useless for our goal.

Negentropy was proposed by \citet{Huber:1985} as a linear invariant index based on the entropy that numerically summarises the departure from the Gaussian distribution. 
The negentropy index for the projected data $\Z$ is defined as follows:
\begin{align}
J(\z)
& = h\big(\phi(\mub_{z},\Sigmab_{z})\big) - h(\z)\nonumber\\
& = \frac{1}{2} \log \big( (2\pi e )^d |\Sigmab_{z}| \big)  + \int \log\big(f(\z)\big) f(\z) \dif\z,
\label{eq:negentropy}
\end{align}
where $h(\phi\big(\mub_{z},\Sigmab_{z})\big)$ is the entropy of the multivariate Normal distribution, $\phi(\mub_{z},\Sigmab_{z})$ is the multivariate Gaussian density of projected data with mean $\mub_{z}$ and covariance matrix $\Sigmab_{z}$, and $h(\z)$ is the entropy of the estimated density for the projected data.

From the definition of the negentropy in the equation \eqref{eq:negentropy}, it can be considered a measure of non-normality \citep{Hyvarinen:etal:2001}.
As shown in \citet[][Appendix A2]{Comon:1994}, negentropy can be rewritten as a Kullback-Leibler divergence, so $J(\z)$ is always  positive, it reaches the minimum value if the density of the projected data is Gaussian, and it is invariant by affine transformations.

Thus, in addition to describing the least normality, negentropy holds all the desirable properties required for a projection index, when visual inspections of multivariate dataset are sought. It is, therefore, suitable for revealing particular structures, which would otherwise be impossible to visualise. In addition, negentropy is a natural choice as a PP index for the purpose of visualising clustering.

The usually unknown density of the projected data, is required to compute negentropy. The linear transformation property of GMMs enables that density to be easily obtained.
However, we do not have a closed formula for the entropy for GMMs, due to the logarithm of the sum of the exponential functions. Even though a Monte Carlo approach would provide the most accurate approximation, it is slow to compute, due to the large sample size needed to achieve reasonable accuracy. More computationally efficient methods attempt to approximate the entropy with closed form solutions and aim to ensure both accuracy and computational efficiency simultaneously.
In the following section, we present different approximation methods available in the literature for the entropy, which are useful to approximate the negentropy index.

\subsubsection{Unscented transformation approximation for negentropy}
\label{ut}
The \emph{Unscented Transformation} (UT) proposed by \citep{Julier:Uhlmann:1996, Goldberger:Aronowitz:2005} is a method for calculating the statistics of a non-linear transformed random variable. Given a $p$-dimensional random vector $\x \sim f(\x)$ and some non-linear transformations $c(\x) : \mathbb{R}^{p}\rightarrow \mathbb{R}$, the method enables $\E_{f}[c(\x)] = \int c(\x)f(\x) \dif\x$ to be approximated. Such approximation is exact if $c(\x)$ is a quadratic function.

In general, for a random variable $\x \in \mathbb{R}^{p}$ such that $\x \sim N(\mub, \Sigmab)$, the approximation is based on the use of a set of $2p$ so-called \emph{sigma-points} to compute the integral. A good choice for these points, which capture the moments of the random variable, is the following:
\begin{equation*}
\left\{
\begin{aligned}
\x_k & = \mub + (\sqrt{p\Sigmab})_k
&&& k = 1, \dots, p \\
\x_{p+k} & = \mub - (\sqrt{p\Sigmab})_k
&&& k = 1, \dots, p
\end{aligned}
\right.
\end{equation*}
where $(\sqrt{\Sigmab})_k$ is the $k$th column of the square root matrix of $\Sigmab$, so $(\sqrt{p\Sigmab})_k = \sqrt{p\lambda_k}\mathbf{u}_k$ with $\lambda_k$ and $\mathbf{u}_k$ respectively the $k$th eigenvalue and eigenvector of $\Sigmab$.
Once the sigma-points have been chosen, the approximation is as follows:
\begin{equation}
\int c(\x) \phi(\x ; \mub, \Sigmab) \dif\x \approx \frac{1}{2p} \sum^{2p}_{k=1} c(\x_k).
\label{UT:Gaussian}
\end{equation}
As in the MC approximation, this method selects some points and computes an average. However, the points are now chosen deterministically and not randomly. Usually, a small number of points is sufficient.
The UT method outlined above can be used to approximate entropy, when the density is a mixture of Gaussian distributions.

Let us consider a model in \eqref{mixtureLinear} and the entropy in \eqref{E}, where $c(\z) = \log f(\z)$. The entropy can be rewritten as follows:
\begin{align*}
  h_{\UT}(\z) & = -\int c(\z)f(\z) \dif\z \\
  & = - \int \log f(\z) \sum^G_{g=1} \pi_g \phi(\z ; \mub_g, \Sigmab_g) \dif\z\\
  & = - \sum^G_{g=1} \pi_g \int \log f(\z)  \phi(\z ; \mub_g, \Sigmab_g) \dif\z \\
  & = - \sum^G_{g=1} \pi_g \E_{\phi_{g}}[\log f(\z)].
\end{align*}
Then, using equation \eqref{UT:Gaussian} we have:
\begin{equation*}
h_{\UT}(\z) = -\frac{1}{2d} \sum^G_{g=1} \pi_g \sum^{2d}_{k=1} \log{f(\z_{gk})},
\end{equation*}
with
\begin{equation*}
\left\{
\begin{aligned}
\z_{gk} & = \mub_g + (\sqrt{d\Sigmab_g})_k
&&& k = 1, \dots, d \\
\z_{g,(d+k)} & = \mub_g - (\sqrt{d\Sigmab_g})_k
&&& k = 1, \dots, d \\
\end{aligned}
\right.
\end{equation*}
Finally, the approximated negentropy obtained with the Unscented Transformation can be written as follows:
\begin{equation*}
J_{\UT}(\z) = \frac{1}{2} \log \big( (2\pi e )^d |\Sigmab_z| \big) - h_{\UT}(\z).
\end{equation*}

\subsubsection{Variational approximation for negentropy}
\label{Dvar}

\citet{Hershey:Olsen:2007} proposed a closed form approximation for the Kullback Leibler (KL) divergence in the case of GMMs. Using this result, we can obtain a closed formula \emph{Variational} approximation for the entropy, which we will denote with VAR.

Assume a GMM $f(\z) = \sum^G_{g=1} \pi_g \phi_g(\z ; \mub_g, \Sigmab_{g})$. Then, by Jensen inequality, the lower bound of minus the entropy of $f(\z)$ can be found as follows:
\begin{align}
- h(\z)
& = \E_{f} \left [ \log \sum_{l=1}^G \pi_{l}\phi_l(\z ; \mub_l, \Sigmab_l) \right ] \nonumber \\
& = \int  \left [ \log \sum_{l=1}^G \pi_{l} \phi_{l}(\z ; \mub_l, \Sigmab_{l})\right ] \times \sum_{g=1}^G \pi_{g}\phi_{g}(\z ; \mub_g, \Sigmab_{g}) \dif\z \nonumber \\
& = \sum_{g=1}^G \pi_{g} \int \left [\log  \sum_{l=1}^G \pi_{l}\phi_{l}(\z ; \mub_l, \Sigmab_{l}) \right ] \times \phi_{g}(\z ; \mub_g, \Sigmab_{g}) \dif\z \nonumber \\
& = \sum_{g=1}^G \pi_{g} \int  \left [ \log  \sum_{l=1}^G \pi_{l} \psi_{lg}\dfrac{\phi_{l}(\z ; \mub_l, \Sigmab_{l})}{\psi_{lg}} \right ]  \times \phi_{g}(\z ; \mub_g, \Sigmab_{g}) \dif\z \nonumber \\
& \geq \sum_{g=1}^G \pi_{g} \sum_{l=1}^G \psi_{lg} \int  \left [ \log \dfrac{\pi_{l}\phi_{l}(\z ; \mub_l, \Sigmab_{l}))}{\psi_{lg}} \right ] \times \phi_{g}(\z ; \mub_g, \Sigmab_{g}) \dif\z,
\label{dvar::entropy}
\end{align}
where $\psi_{lg}$ is a variational parameter, such that $\sum_{l=1}^{G}\psi_{lg} = 1$. 
Then, by maximising the right hand of the equation \eqref{dvar::entropy} with respect to $\psi_{lg}$, the VAR approximation for the GMMs entropy can be shown to be the following \citep{Hershey:Olsen:2007}:
\begin{equation*}
h_{\VAR}(\z) = \sum_{g = 1}^{G} \pi_{g} \log \sum_{l= 1}^{G} \pi_{l} 
               \exp\{D\big( \phi_g (\z ; \mub_g, \Sigmab_{g}) || \phi_{l}(\z ; \mub_l, \Sigmab_{l}) \big)\} 
               - \sum_{g =1}^{G} \pi_{g} h\big(\phi_{g}(\z ; \mub_g, \Sigmab_{g})\big),
\end{equation*}
where $D\big( \phi_g (\z ; \mub_g, \Sigmab_{g}) || \phi_{l}(\z ; \mub_l, \Sigmab_{l}) \big)$ is the KL-divergence between the density functions 
$\phi_g (\z ; \mub_g, \Sigmab_{g})$ and $\phi_{l}(\z ; \mub_l, \Sigmab_{l})$, and
$h\big(\phi_{g}(\z ; \mub_g, \Sigmab_{g})\big)$ is the entropy of the $g$th component of the mixture.

By using the above variational approximation for the entropy, the closed formula approximation for the negentropy is the following:
\begin{equation*}
J_{\VAR}(\z) = \frac{1}{2} \log \big( (2\pi e )^d |\Sigmab_z| \big) - h_{\VAR}(\z).
\end{equation*}

\subsubsection{Second order Taylor expansion approximation for GMM negentropy}
\label{H}

\citet{Huber:etal:2008} proposed a \emph{second order Taylor expansion} to approximate the entropy for GMMs. For the remainder of the work we shall refer to this approximation as SOTE.
The approximated entropy for the model \eqref{mixtureLinear} can be written as follows:
\begin{equation*}
h_{\SOTE}(\z) = h_0(\z) - \sum^{G}_{g=1} \dfrac{\pi_g}{2} F(\mub_g) \odot \Sigmab_g,
\end{equation*}
where $h_0(\z) = -\sum^{G}_{g=1} \pi_g \log \phi(\mub_g ; \mub_g,\Sigmab_g)$ is the first order expansion of the entropy around the mean vector $\mub_g$, $\odot$ is the so-called matrix contradiction operator, so that for the two matrices $\A \in \mathbb{R}^{n \times m}$ and $\B \in \mathbb{R}^{n \times m}$, $\A \odot \B = \sum_{i=1}^{n}\sum_{j=1}^{m} a_{ij}b_{ij}$, and
\begin{multline*}
F(\x) = \dfrac{1}{f(\x)} \sum^{G}_{g=1} \pi_g \Sigmab^{-1}_g \left( \dfrac{1}{f(\x)}(\mub_g -\x) (\nabla f(\x))\T + (\mub_g -\x)\left(\Sigmab^{-1}_g(\x - \mub_g)\right)\T - \I \right) \\ \times \phi(\x ;\mub_g, \Sigmab_g),
\end{multline*}
for a generic vector $\x$, where $\nabla f(\x)$ is the gradient of the mixture model with respect to the data.
Thus, the second order Taylor approximation for the entropy generates the following negentropy:
\begin{equation*}
J_{\SOTE}(\z) = \frac{1}{2} \log \big( (2\pi e )^d |\Sigmab| \big) - h_{\SOTE}(\z).
\end{equation*}

\subsection{Maximisation of GMMs negentropy}
\label{sec:maxgmmnegent}

GAs are stochastic optimisation methods belonging to the class of evolutionary algorithms, which use the theory of natural selection and biological evolution to seek the optimal solution. Introduced by \citet{Holland:1975} to study the principle of adaptive system with bit strings representation, over the years GAs have been applied to optimisation problems with continuous variable representation \citep{Goldberg:Holland:1988}.
The algorithm does not need the computation of derivatives and can, therefore, be applied to discrete and continuous problems without any concerns for the properties of the objective function.

The idea behind this type of algorithm is that, given a population, only the fittest individual survives at the pressure of the environment (natural selection). Each individual of the population is a candidate solution for a maximisation problem, and fitness is the value of the objective function to maximise. If only the fittest individual survives, it generates an increase in the value of the objective function.

The solution space, where GAs operate, is not the same space as that of the original problem. Instead it is a different space called the representations space.
The decision variables in the GAs are encoded using an encoding function. The encoded individual is the candidate solution for the GAs which, when decoded, represents the candidate solution of the true maximisation problem.
The encoding function modifies the original problem into a more abstract, but feasible solutions space, where GAs works, allowing it to be moved from the GAs solutions space to the original solutions space of the problem. Let $B$ be the problem solution space and $\Theta$ the GA solution space, then the encoding function $E:B \rightarrow \Theta$ is any one-to-one function that maps the solutions of the problem to GA solutions/individuals. The inverse relation is called the decoding function \citep{Dumitrescu:2000}.

In the proposed PP procedure, a sine-cosine transformation proposed by \citet{Baragona:Battaglia:Poli:2011} is employed to obtain a finite parameters space for GAs.
Let $\b_{j} = (b_{1j}, \dots , b_{pj})$ with $j = 1, \dots, d$ denote the generic column of the basis matrix $\B$. The proposed transformation is as follows:
\begin{align*}
b_{1j} & = \sin(\theta_{1j})\sin(\theta_{2j}) \dots \sin(\theta_{(p-2)j})\sin(\phi_j) \\
b_{2j} &= \sin(\theta_{1j})\sin(\theta_{2j}) \dots \sin(\theta_{(p-2)j})\cos(\phi_j) \\
b_{3j} & = \sin(\theta_{1j})\sin(\theta_{2j}) \dots \cos(\theta_{(p-2)j}) \\
b_{4j} & = \sin(\theta_{1j})\sin(\theta_{2j}) \dots \cos(\theta_{(p-3)j}) \\
b_{5j} & = \sin(\theta_{1j})\sin(\theta_{2j}) \dots \cos(\theta_{(p-4)j}) \\
\vdots \\
b_{pj} & = \cos(\theta_{1j}),
\end{align*}
where $0 \leq \phi_j \leq 2\pi$ and $0 \leq \theta_{ij} \leq \pi$, for $j = 1, \dots d$ and $i = 1, \dots, p$. The algorithm then explores every feasible solution in the parameter space formed by $\Theta = \{ \theta_{1j}, \dots, \theta_{(p-2)j}, \phi_j ; \forall\ j = 1, \dots d  \}$.

The basis matrix $\B$ is obtained by applying the decoding function to the GA solution:
\begin{equation}
\label{encode:B}
\phi_{1},\theta_{11},\dots,\theta_{(p-2)1},\phi_{2},\theta_{21},\dots,\theta_{(p-2)1}, \dots, \phi_{d},\theta_{1d},\dots,\theta_{(p-2)d},
\end{equation}
where the first $p-1$ values are used to decode the first column vector of $\B$, then the second $p-1$ values are used to decode the second column vector of $\B$, and so on. 
The entire GA solution in \eqref{encode:B} represents a single individual of the GA population.

New solutions, called offsprings, are generated by stochastically applying genetic operators, such as selection, crossover and mutation, to the individuals of the population.
Parents are selected based on their fitness, and then recombined to generate new offsprings. Mutation perturbs some values of offsprings, to slightly modify the new solutions. Offsprings survive their parents if their fitness values are greater than the parents' fitness, and then substituted in the populations. An elitism strategy may also be applied by preserving the best fitted individual(s) in the next iterations.
The procedure is repeated until some convergence criterion is reached, and then the individual/solution is returned. In our context, the individual/solution is the decoded/encoded basis, and the fitness is the approximated negentropy index.

\subsection{A note on computing time}

To investigate the runtime of the proposed approach, we designed a simulation study in which bivariate data were simulated from a mixture of three Gaussian components having the same covariance structure 
$\Sigmab_g = 
\left[\begin{smallmatrix}
 0.1 & 0 \\ 
 0   & 0.1 
 \end{smallmatrix} \right]$
for $g=1,2,3$, and component means located at the vertices of an equilateral triangle, i.e. 
$\mub_1 = [-1, -1]\T$, $\mub_2 = [0, 1]\T$, and $\mub_3 = [1, -1]\T$. 

\begin{figure}[!htb]
  \centering
  \includegraphics[width=0.8\textwidth]{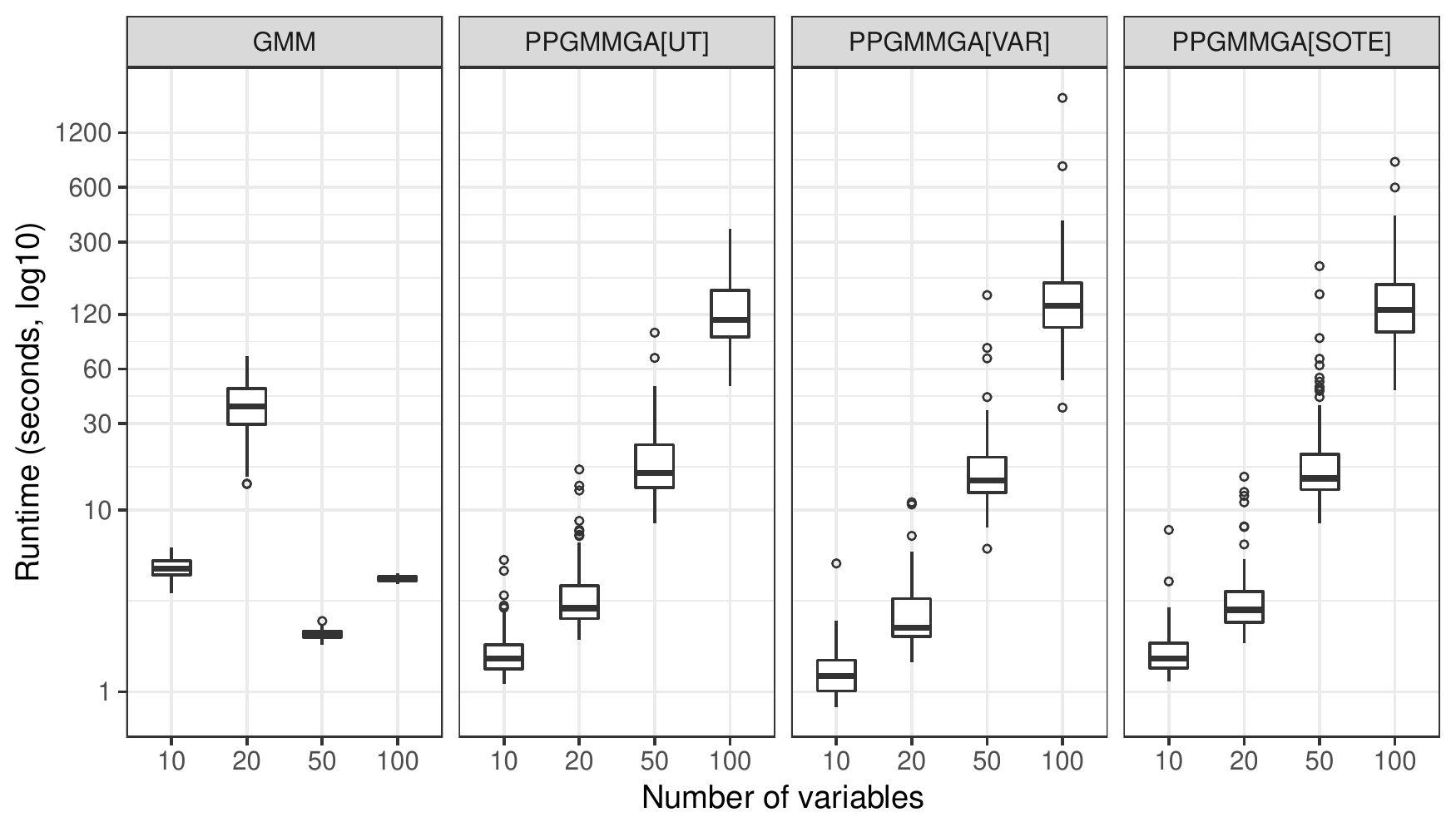}
  \caption{Box-plots of runtimes (expressed in seconds on a $\log_{10}$-scale) versus the number of variables. Data were generated from a three components bivariate Gaussian mixture with equal covariance matrices, whereas the remaining variables were generated from independent standard Gaussians. 
The runtimes refer to the time required for fitting GMMs, and PPGMMGA using the three different negentropy approximations discussed in Section~\ref{sec:negentprojind} (i.e. PPGMMGA[UT] for the Unscented Transformation, PPGMMGA[VAR] for the VARiational, and PPGMMGA[SOTE] for the Second Order Taylor Expansion). The overall execution time of the PPGMMGA procedure can be obtained by summing the time required for fitting the GMM, and the time spent on maximising one of the negentropy approximations.}
  \label{fig:runtime}
\end{figure}

The remaining variables were generated from independent standard Gaussians. We drew samples of size $n=500$ for $p \in \{10, 20, 50, 100\}$ variables, and we recorded the runtimes for the density estimation step using GMMs, and for the projection pursuit step using PPGMMGA with the three negentropy approximations discussed in Section~\ref{sec:negentprojind}.
The GMM density estimation step was performed using the \pkg{mclust} R package \citep{Scrucca:etal:2016}, using BIC to select both the number of mixture components and the components covariance structure.
Simulations were carried out on a iMac with 4 cores i5 Intel CPU running at 2.8 GHz and with 16GB of RAM.
Figure~\ref{fig:runtime} shows the box-plots of computing times (in seconds on a $\log_{10}$-scale) for 100 replications. Note that, the overall runtime is given by the sum of the time needed for fitting GMMs plus the time required for running PPGMMGA with one of the negentropy approximations available.

The runtime needed to execute the PPGMMGA procedure is comparable among the three negentropy approximations, ranging from few seconds when $p$ is small or moderate, to few minutes when $p$ is the largest. The GMMs computing time appears odd at first sight, nevertheless it can be easily explained. 
By default, all models with up to 9 components and 14 different covariance structures are fitted, for a total of 126 estimated models \citep[see ][Table~3]{Scrucca:etal:2016}. 
This is the case when $p=10$ or $p=20$. However, for $p=50$ or $p=100$ only a small subset of covariance structures can be estimated, i.e. the most parsimonious ones, and this causes an acceleration of the density estimation process.
  
\section{Data analysis examples}
\label{sec:examples}

In this section, we present the results obtained using the proposed PPGMMGA  algorithm with different approximations on both real and simulated datasets.

The approximations used to estimate the negentropy for GMMs in closed form are all affected by an approximation error. This means the maximised negentropy values of the different methods cannot be compared. Consequently, a comparable measure needs to be introduced to compare the different PPGMMGA approximations.
Since the MC approximation of the negentropy is the only one guaranteed to converge to the true value of the negentropy, the MC approximation is used to compare the results obtained from the different procedures. For each proposed approximation, the PPGMMGA procedure is performed, and the final, estimated, orthogonal basis is retained.
Let $\hat{\B}_{a}$ be the basis of the projection subspace, estimated using the approximation $a$, then the Monte Carlo negentropy is computed as follows:
\begin{equation*}
J_{\MC,a}(\z) = \frac{1}{2} \log \big( (2\pi e )^d |\Sigmab| \big) - h_{\MC}(\hat{\B}_{a}\T \x).
\end{equation*}
A relative measure can be derived to compare the approximations in terms of how well they approximate the MC version of the negentropy;
\begin{equation*}
J_{a}^{\Rel} = \frac{J_{a}(\z)}{J_{\MC,a}(\z)}.
\end{equation*}
Values greater than 1 indicate the considered approximation overestimates the true value of the negentropy, whereas values smaller than 1 indicate that the approximation tends to underestimate the true negentropy. Values around 1 indicate a good accuracy of the considered approximation. $100,000$ MC samples were employed to estimate the true negentropy.

The projections obtained using two different PPGMMGA approximations were compared by measuring the distance between two subspaces \citep{Li:Zha:Chiaromonte:2005}.
Let $S(\B_{1})$ and $S(\B_{2})$ be two $d$-dimensional subspaces of $\mathbb{R}^p$, spanned respectively by $\B_{1}$ and $\B_{2}$. In addition, let $P_{S(\B_{1})}$ and $P_{S(\B_{2})}$ be the orthogonal projections onto $S(\B_{1})$  and  $S(\B_{2})$. Then, the distance between these two subspaces is as follows:
\begin{align}
\Delta(\B_{1},\B_{2})
& = || P_{S(\B_{1})} - P_{S(\B_{2})} || \nonumber \\
& = || \B_{1}(\B_{1}\T\B_{1})^{-1}\B_{1}\T - \B_{2}(\B_{2}\T\B_{2})^{-1}\B_{2}\T ||,
\label{angle}
\end{align}
where $||.||$ is the spectral Euclidean norm.  It can be shown that $\Delta(\B_{1},\B_{2}) = \arcsin \alpha \in [0,1]$, i.e. the maximal angle $\alpha$ between the two subspaces of $\mathbb{R}^p$. This angle can be expressed in degree, i.e.  $\Delta\degree (\B_{1},\B_{2}) = \arcsin (\Delta(\B_{1},\B_{2})) \times 180 / \pi \in [0\degree, 90\degree]$, where $\Delta\degree (\B_{1},\B_{2}) = 0\degree$ for the equivalent subspaces and $\Delta\degree (\B_{1},\B_{2}) = 90\degree$ for completely orthogonal subspaces. This conversion is used in the results.

GMM density estimation is performed using the \pkg{mclust} R package \citep{Scrucca:etal:2016}. The function \code{densityMclust()} provides estimates for the parameters of a Gaussian mixture model obtained via EM algorithm. The BIC criterion is employed to select the model, i.e. either the number of mixture components and the components covariance structure.

The \pkg{GA} package of R \citep{Scrucca:2013, Scrucca:2017} is used to implement the GA optimisation.
The package implements Hybrid-GAs \citep{Scrucca:2017, Eiben:Smith:2003}, a type of algorithm that incorporates local search algorithms, e.g. quasi-Newton, conjugate-gradient, and simplex based algorithms, to combine the properties of GAs with a local maximiser. Thus, hybridisation is used in the proposed PP procedure, and the local optimiser employed is a modification of the Newton algorithm proposed by \citet{Byrd:etal:1995}, which allows lower and upper bounds for the variables.

As genetic operators, we employ the proportional selection with fitness linear scaling \citep{Dumitrescu:2000, Simon:2013} to select two parents, a local arithmetic crossover \citep{Simon:2013} to recombine the two selected parents, and uniform perturbation \citep{Simon:2013} as a mutation strategy.
Selection, crossover and mutation are all implemented in the \pkg{GA} package. Genetic operators are applied stochastically at each iteration, so they must be specified before beginning the procedure. We set the initial population size at $100$ individuals, the probability of crossover at $0.8$, the probability of mutation at $0.1$, and the local search probability at $0.05$. An elitism strategy was also employed, so the best individual is automatically included in the next generation.

Principal Component Analysis (PCA) is a popular algorithm for dimension reduction. It is well known that, if a large variance is considered in defining a projection index, then PCA can be seen as a special case of projection pursuit \citep{Jolliffe:2002}. However, \citet{Chang:1983} showed that using the first PCA directions might fail to reveal any clustering structure. Despite this fact, we included PCA in the analyses for the purpose of comparison. 
In addition, Independent Component Analysis \citep[ICA;][]{Hyvarinen:etal:2001} was also included in such a comparison. ICA is a blind source separation methodology which aims at extracting independent non-Gaussian signals from multivariate data. Among the many proposals available in the literature, fastICA is an efficient and popular algorithm for maximising an approximation of the negentropy \citep{Hyvarinen:1999}, and an implementation is available in the R package \code{fastICA} \citep{fastICA}.

The R package \code{ppgmmga} implementing the proposed methodology is used in the following data analysis examples. Data have been always centred and scaled unless otherwise specified. A script file containing R code for replicating the examples is also provided in the Supplemental Materials. 

\subsection{Waveform data}

This artificial, three-class data with 21 variables is often used in statistical and machine learning literature and it is considered a difficult pattern recognition problem \citep{Breiman:etal:1984, Hastie:Tibshirani:1996}. Three shifted triangular waveforms are defined as follows:
\begin{equation*}
w_1(j) = \max(6 - |j-11|,0), \quad w_2(j) = w_1(j-4), \quad w_3(j) = w_1(j+4),
\end{equation*}
for $j = 1,\dots,21$. Each variable $X_j$ is generated within each class $g$ as a random convex combination of two basic waveforms with added noise:
\begin{equation*}
X_j =
\begin{cases}
u_1w_1(j) + (1-u_1)w_2(j) + \epsilon_j & \quad\text{if $g=1$} \\
u_2w_2(j) + (1-u_2)w_3(j) + \epsilon_j & \quad\text{if $g=2$} \\
u_3w_3(j) + (1-u_3)w_1(j) + \epsilon_j & \quad\text{if $g=3$}
\end{cases}\;,
\end{equation*}
for $j=1,2,\dots,21$, where $w_{h} = (w_{h}(1),\dots,w_{h}(21))\T$ for $h = 1,2,3,(u_{1},u_{2},u_{3})$ are independent random variables uniformly distributed on $[0,1]$, and $\epsilon_{j}$ is a random noise following a standard normal distribution.
A triangular shape of the projected data is expected with each group forming one side of the triangle. A random sample of size $n=400$ was drawn, and the data were only centred before analysis; no scaling was applied because the features are expressed in the same unit of measurement.

\begin{figure}[!htb]
  \centering
  \includegraphics[width=\textwidth]{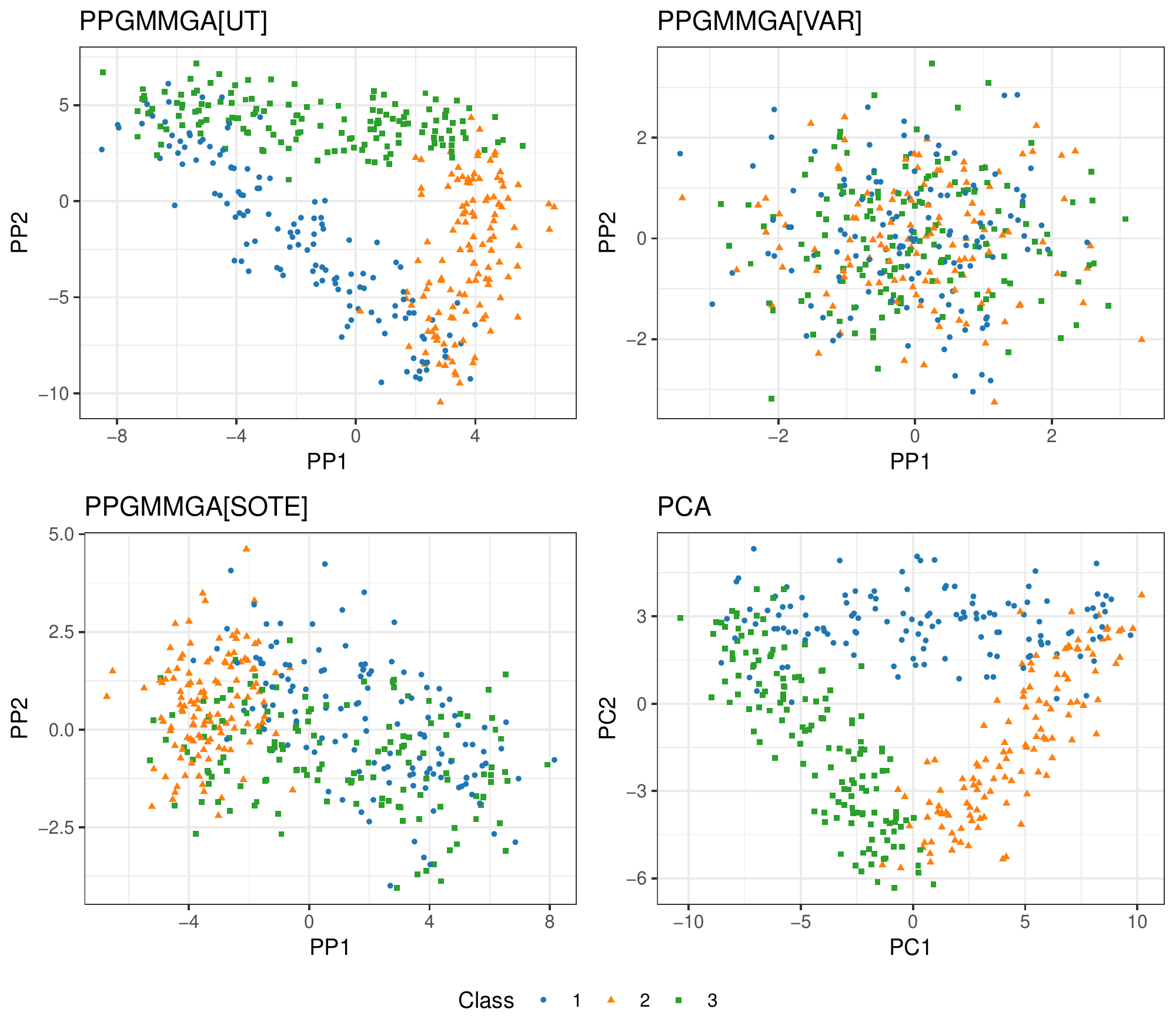}
  \caption{Scatterplots of Waveform data projected using the optimal 2-dimensional PPGMMGA with different negentropy approximations (PPGMMGA[UT] for the Unscented Transformation, PPGMMGA[VAR] for the VARiational, and PPGMMGA[SOTE] for the Second Order Taylor Expansion), and the first two principal components (PCA). Points are marked according to the original class they belong to. Only PPGMMGA[UT] (top-left panel) and PCA (bottom-right) are able to show the original three-group triangular structure.}
  \label{fig:wave}
\end{figure}

Table~\ref{tab:waive} shows the results obtained by applying the PPGMMGA procedure with $d=2$, using different negentropy approximations for a comparison of subspaces. Figure~\ref{fig:wave} shows the corresponding 2D visualisations.

\begin{table}[!htb]
  \centering
  \begin{tabular}{rccccc}
  \toprule
  PP index          & UT & VAR & SOTE & PCA & ICA \\
  \cline{2-6}
  Negentropy        & 1.0025 & 0.2333 & 0.4216 & --- & --- \\
  MC negentropy     & 1.0210 & 0.2389 & 0.4332 & 1.0285 & 1.0313 \\
  Relative accuracy & 0.9818 & 0.9764 & 0.9732 & --- & --- \\
  \midrule
  Angle  & UT & VAR & SOTE & PCA & ICA \\
  \cline{2-6}
  UT & 0.00 & 89.02 & 80.48 & 9.03 & 9.03 \\
  VAR &  & 0.00 & 85.48 & 89.49 & 89.49 \\
  SOTE &  &  & 0.00 & 80.51 & 80.51 \\
  \bottomrule
  \end{tabular}
  \caption{Maximised negentropy obtained using different 2D PPGMMGA index approximations, MC negentropy and the corresponding relative accuracy, for the Waveform data. PCA and ICA are also included for comparison. The bottom part of the table shows the angles (in degrees) between the estimated subspaces.}
  \label{tab:waive}
\end{table}

The UT approximation yields an estimate of the projection subspace which outperforms the other two methods, and clearly shows the underlying structure in the data.
All the approximations are reasonably close to the MC negentropy. However, both VAR and SOTE approaches give solutions far from the optimal subspace.
Overall, the procedure proposed with the UT approximation appears to be able to recover the main structure of the data.
Lastly, it should be noted that the estimated basis and the corresponding negentropy for both PCA and ICA are very close to PPGMMGA with UT approximation in this case, although this is not guaranteed to occur in general, as the following examples will show.

\subsection{Australian crabs data}

The Australian crabs data contains physical measurements on 200 Leptograpsus crabs in Western Australia \citep{Campbell:Mahon:1974} and can be found in the \pkg{MASS} R package \citep{MASS}.
There are five measurements for each crab: frontal lobe size (\code{FL}), rear width (\code{RW}), carapace length (\code{CL}), carapace width (\code{CW}), and body depth (\code{BD}). 
Furthermore, crabs can be classified according to their colour (blue and orange) and gender, giving four groups. Fifty specimens are available for each combination of colour and gender.

As reported in Table~\ref{tab:crabs}, the UT approximation achieves the largest negentropy. By comparing its value against the MC negentropy, the UT approximation is also the most accurate, whereas the VAR and SOTE approximations largely underestimate and overestimate, respectively, the negentropy obtained with Monte Carlo. Looking at the angles between the estimated projection subspaces, the UT solution appears to be quite different from the others, including the PCA and ICA projections.

\begin{table}[!htb]
  \centering
  \begin{tabular}{rccccc}
  \toprule
  PP index & UT & VAR & SOTE & PCA & ICA \\
  \cline{2-6}
  Negentropy        & 0.6001 & 0.2716 & 0.5684 & --- & --- \\
  MC negentropy     & 0.6078 & 0.4575 & 0.4905 & 0.1933 & 0.1898 \\
  Relative accuracy & 0.9875 & 0.5937 & 1.1589 & --- & --- \\
  \midrule
  Angle & UT & VAR & SOTE & PCA & ICA \\
  \cline{2-6}
  UT   & 0.00 & 75.99 & 66.38 & 88.11 & 88.11 \\
  VAR  &  & 0.00 & 85.25 & 88.20 & 88.20 \\
  SOTE &  &  & 0.00 & 88.35 & 88.35 \\
  \bottomrule
  \end{tabular}
  \caption{Maximised negentropy obtained using different 2D PPGMMGA index approximations, MC negentropy and the corresponding relative accuracy, for the Australian crabs data. PCA and ICA are also included for comparison. The bottom part of the table shows the angles (in degrees) between the estimated subspaces.}
  \label{tab:crabs}
\end{table}

\begin{figure}[!htb]
  \centering
  \includegraphics[width=0.9\textwidth]{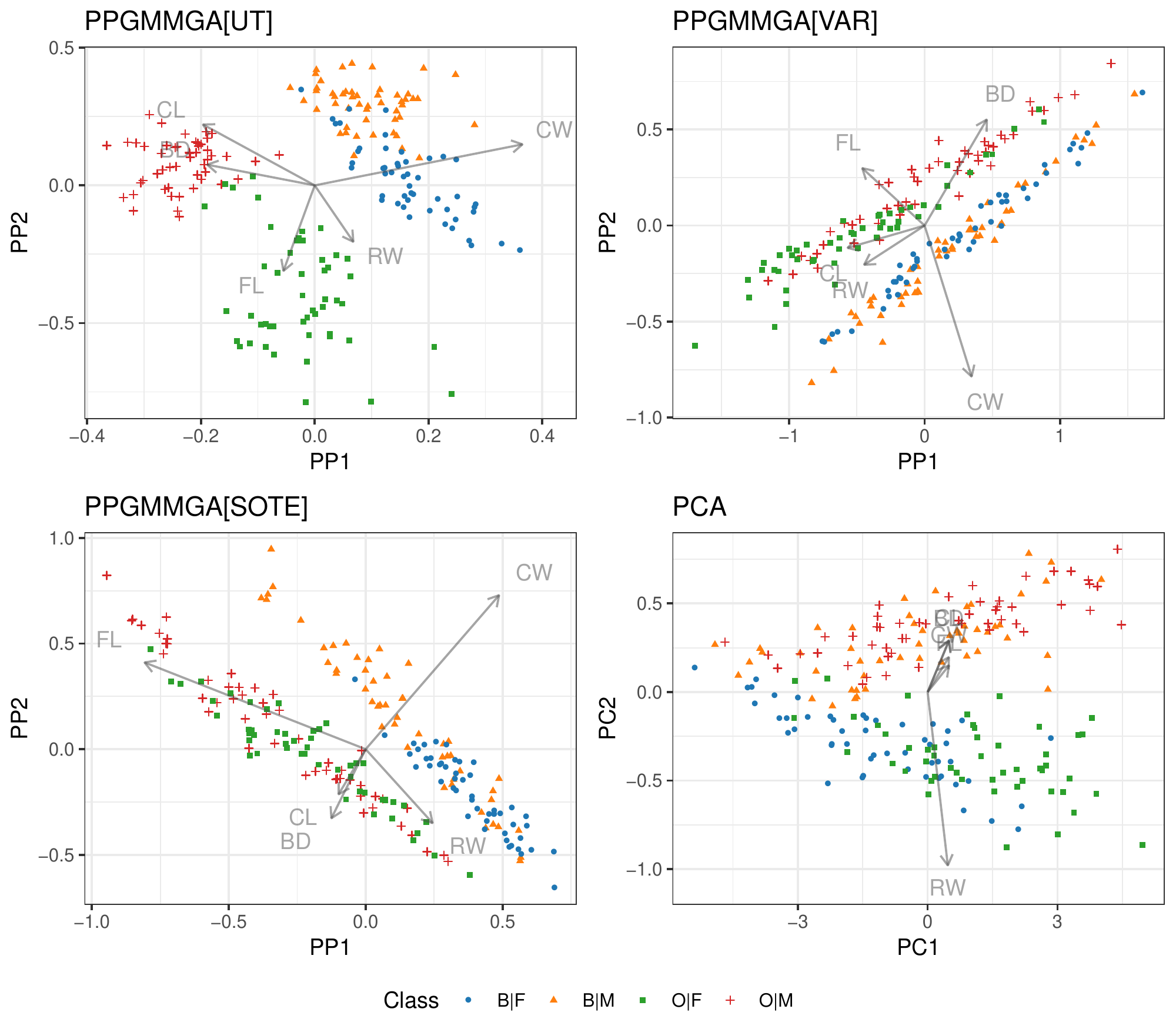}
    \caption{Scatterplots of Australian crabs data projected using the optimal 2-dimensional PPGMMGA with different negentropy approximations (PPGMMGA[UT] for the Unscented Transformation, PPGMMGA[VAR] for the VARiational, and PPGMMGA[SOTE] for the Second Order Taylor Expansion), and the first two principal components (PCA). Points are marked according to the original classification: B|F indicates blue female crabs, B|M indicates blue male crabs, O|F indicates female orange crabs and O|M indicates orange male crabs.
Arrows indicate the biplot vectors corresponding to the original features, i.e. frontal lobe size (\code{FL}), rear width (\code{RW}), carapace length (\code{CL}), carapace width (\code{CW}), and body depth (\code{BD}). 
The PPGMMGA[UT] projection (top-left panel) is the only able to separate the crabs by colour along the carapace width (\code{CW}) and by gender using the other features.} 
  \label{fig:crabs}
\end{figure}

Figure~\ref{fig:crabs} shows the estimated $2$-dimensional PPGMMGA projections. Clearly, the UT approximation is able to separate the data in two different groups corresponding to the colour of the crabs along the \code{CW} direction, whereas the remaining variables allows the crabs to be separated by gender. 
Both the VAR and SOTE solutions manage to separate crabs with respect to colour, but not on gender. 
In contrast, the PCA solution (as well as the ICA, although not shown) presents a V-shape with the two arms that allow to distinguish the crabs according to the gender, but fails to reveal any separation based on colour.

\subsection{Coffee data}

Coffee data \citep{Streuli:1973} provides the chemical composition of two varieties of coffee (Arabica, Robusta) for 43 samples collected from 29 countries around the world. Twelve measurements were considered for each sample. The dataset is available in the \pkg{pgmm} R package \citep{pgmm}.

A 1-dimensional PPGMMGA procedure was performed and the results are reported in Table~\ref{tab:coffee}. In this case, all the approximations essentially achieve the same value of negentropy, and show very good accuracy when compared to the Monte Carlo entropy. As a result, the angles between the estimated subspaces are close to zero, hence spanning almost the same projection subspace.

\begin{table}[!htb]
  \centering
  \begin{tabular}{rccccc}
  \toprule
  PP index & UT & VAR & SOTE & PCA & ICA \\
  \cline{2-6}
  Negentropy        & 1.0732 & 1.0730 & 1.0732 & --- & --- \\
  MC negentropy     & 1.0729 & 1.0740 & 1.0753 & 0.7731 & 0.7682 \\
  Relative accuracy & 1.0003 & 0.9991 & 0.9981 & --- & --- \\
  \midrule
  Angle  & UT & VAR & SOTE & PCA & ICA \\
  \cline{2-6}
  UT   & 0.00 & 0.25 & 0.01 & 34.12 & 34.12 \\
  VAR  &  & 0.00 & 0.25 & 34.17 & 34.17 \\
  SOTE &  &  & 0.00 &  34.12 & 34.12 \\
  \bottomrule
  \end{tabular}
  \caption{Maximised negentropy obtained using different 1D PPGMMGA index approximations, MC negentropy and the corresponding relative accuracy, for the Coffee data. PCA and ICA are also included for comparison. The bottom part of the table shows the angles (in degrees) between the estimated subspaces.}
  \label{tab:coffee}
\end{table}

\begin{figure}[!htb]
  \centering
  \includegraphics[width=\textwidth]{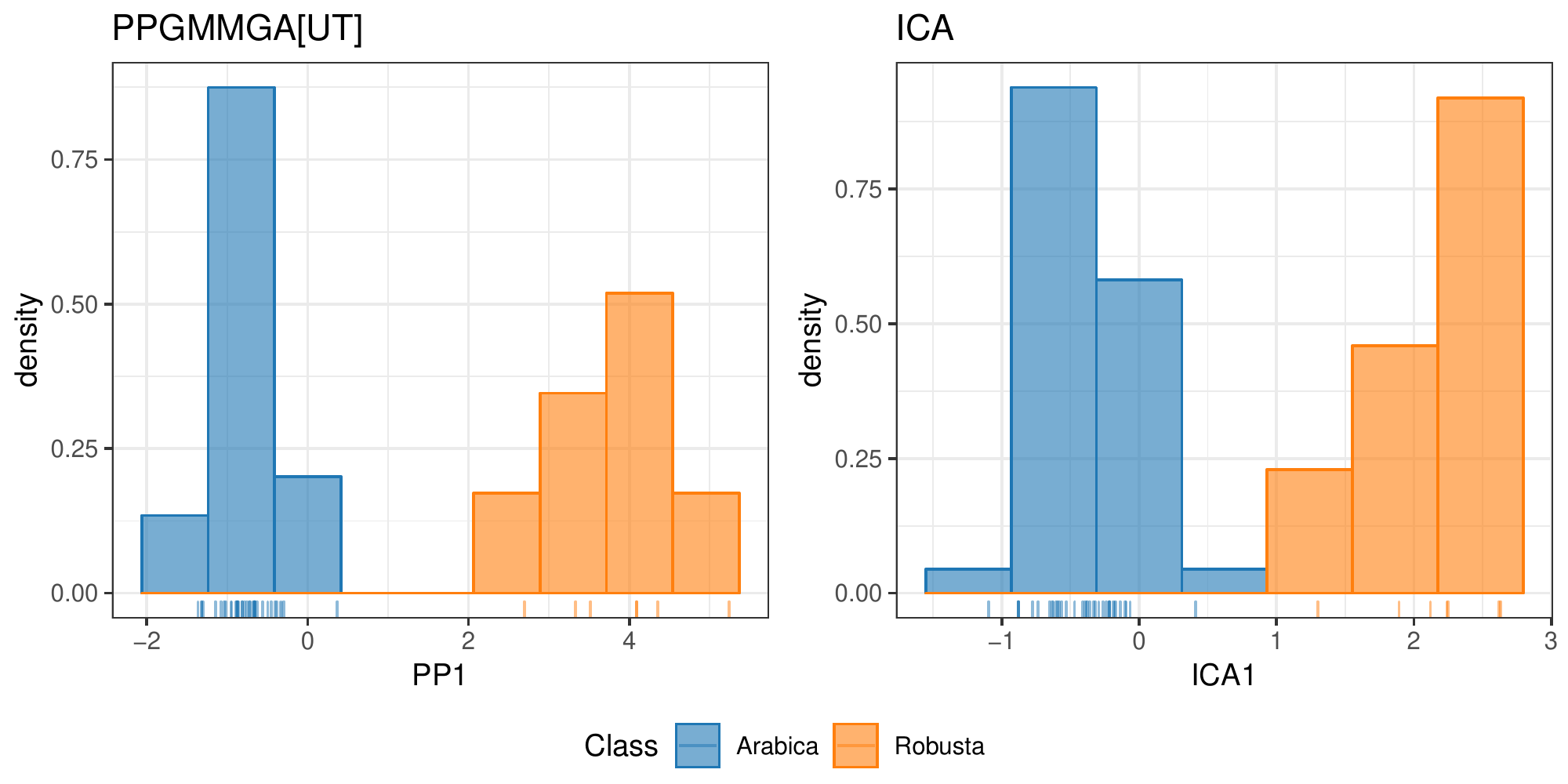}
    \caption{  
  Histograms of optimal 1-dimensional projections obtained using PPGMMGA with the Unscented Transformation (PPGMMGA[UT]; left panel) and Independent Component Analysis (ICA; right panel) for the Coffee data. Since all the PPGMMGA approximations achieve essentially the same results, only one projection is reported, together with ICA for comparison. Bins are coloured according to the coffee variety, Arabica and Robusta. Both methods are able to separate the coffee varieties using a 1-dimensional projection, with a much larger separation shown in the direction estimated by PPGMMGA.}
  \label{fig:coffee1}
\end{figure}

Since the choice of the approximation method employed in the PPGMMGA procedure appears to be irrelevant for the coffee data, only a single 1-dimensional projection is shown in the left-hand panel of Figure~\ref{fig:coffee1}. For comparison, the first ICA projection is also provided. It is interesting to note that, as anticipated by the larger value of the negentropy, the two varieties of coffee are much clearly separated along the single direction estimated by PPGMMGA.

\begin{figure}[!htb]
  \centering
  \includegraphics[width=0.9\textwidth]{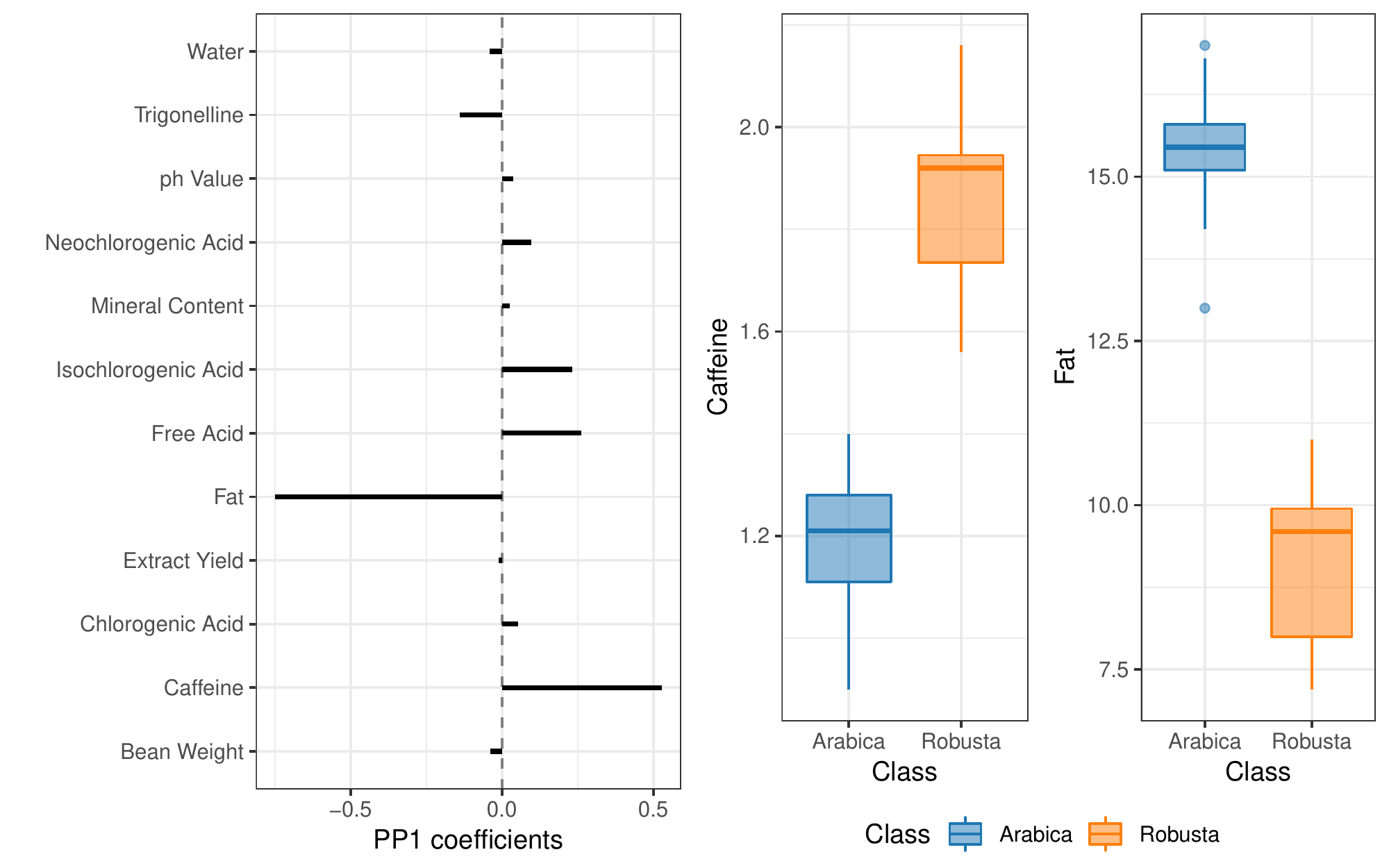}
  \caption{The left panel shows the estimated coefficients that define the 1D PPGMMGA projection, indicating that the features that most contribute to the estimated direction are \code{Fat} and \code{Caffeine}. 
The right panel provides the box-plots of such features conditioning on the coffee varieties, suggesting that high (low) \code{Caffeine} and low (high) \code{Fat} characterise the Robusta (Arabica) variety.}
  \label{fig:coffee2}
\end{figure}

The left panel of Figure~\ref{fig:coffee2} shows the estimated coefficients for the 1D PPGMMGA projection. Based on this chart, we can argue that two features, namely \code{Caffeine} and \code{Fat},  mainly define the estimated direction.
The right panel of Figure~\ref{fig:coffee2}, which contains the conditional box-plots for these two characteristics, suggests that the Robusta variety of coffee has a high content of caffeine but a low content of fat, whereas the Arabica variety has a low content of caffeine but a high content of fat.
The identified features provide an easy interpretation of the main aspects that differentiate coffee varieties, and they could be the only ones required if we wish to discriminate between the two types of coffee.

\subsection{Australian Institute of Sports (AIS) data}

The dataset contains eleven biometric observations, red cell count (\code{rcc}), white cell count (\code{wcc}), hematocrit (\code{Hc}), hemoglobin (\code{Hg}), plasma ferritin (\code{Fe}), body mass index (\code{bmi}), sum of skin folds (\code{ssf}), body fat percentage (\code{Bfat}), lean body mass (\code{lbm}), height (\code{Ht}) and weight (\code{Wt}) on 102 male Australian athletes and 100 female Australian athletes, collected at the Australian Institute of Sport \citep{Cook:Weisberg:1994}.
The AIS data, included in the \pkg{dr} R package \citep{dr}, is used to test the PPGMMGA procedure in the case where the underlying structure of the data is not a Gaussian mixture. Thus, different distributions could be used to better model such data \citep{Azzalini:DallaValle:1996, Arnold:Beaver:2000,
Morris:McNicholas:2013}. 

Table \ref{tab:ais} shows the results for the 1- and 2-dimensional PPGMMGA. In the 1-dimensional case, all the approximations were very inaccurate, especially for the SOTE method. Moving on to the 2-dimensional PPGMMGA, the negentropy approximations increased for all the methods, and the accuracy largely improved, with the exception of SOTE which appeared again to be very misleading. Negentropy for the PCA is substantially lower for both 1- and 2-dimensional cases, with the resulting bases which are almost orthogonal to those estimated by PPGMMGA. Results for ICA are not reported because they were essentially equivalent to those of PCA.

\begin{table}[!htb]
  \centering\small
  \begin{tabular}{rlcccclcccc}
    \toprule
    && \multicolumn{4}{c}{1D} && \multicolumn{4}{c}{2D} \\
    \midrule
    PP index && UT & VAR & SOTE & PCA && UT & VAR & SOTE & PCA \\ 
    \cline{3-6}\cline{8-11}
    Negentropy        && 0.2716 & 0.2283 & 0.5529 & --- 
                      && 0.9187 & 0.6639 & 1.0429 & --- \\
    MC negentropy     && 0.2236 & 0.3071 & 0.1007 & 0.1249
                      && 0.9113 & 0.8600 & 0.2350 & 0.2703 \\
    Relative accuracy && 1.2147 & 0.7435 & 5.4892 & --- 
                      && 1.0081 & 0.7720 & 4.4372 & --- \\
    \midrule
    Angle  && UT & VAR & SOTE & PCA && UT & VAR & SOTE & PCA \\
    \cline{3-6}\cline{8-11}
    UT   && 0.00 & 34.80 & 27.81 & 81.53
         && 0.00 & 12.19 & 55.23 & 88.65 \\
    VAR  &&      &  0.00 & 44.15 & 86.59
         &&      &  0.00 & 59.97 & 88.51 \\    
    SOTE &&      &       &  0.00 & 87.06
         &&      &       &  0.00 & 89.65 \\    
    \toprule
  \end{tabular}
  \caption{Maximised negentropy obtained using different 1D and 2D PP index approximations, MC negentropy and the corresponding relative accuracy, for the AIS data. The PCA is also included for comparison. The bottom part of the table shows the angles (in degrees) between the estimated subspaces.}
  \label{tab:ais}
\end{table}

\begin{figure}[!htb]
  \centering
  \includegraphics[width=\textwidth]{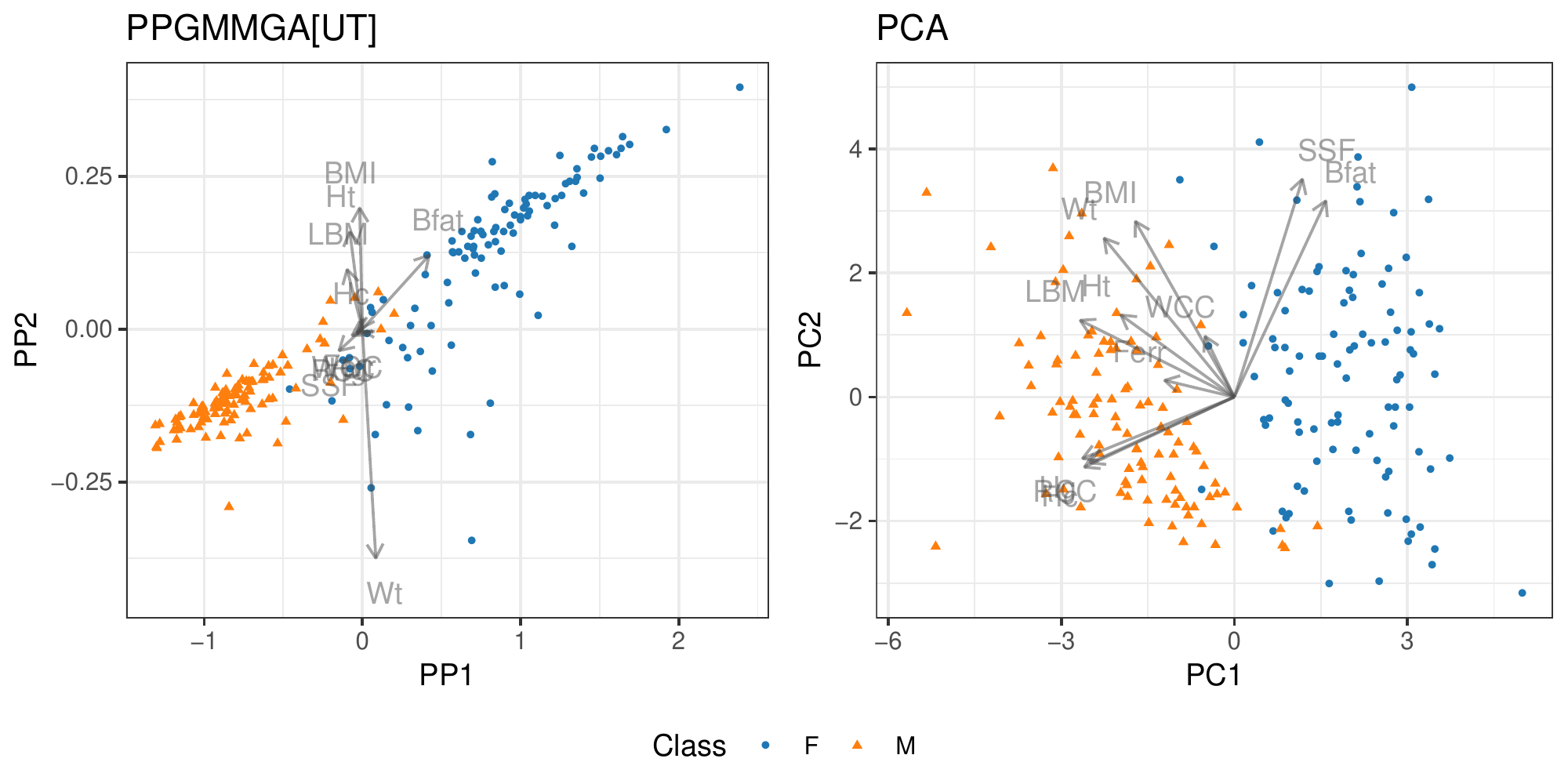}
  \caption{Scatterplots of AIS data projected using the optimal 2-dimensional PPGMMGA using Unscented Transformation (PPGMMGA[UT]; left panel), and the first two principal components (PCA; right panel). Points are marked according to the athletes' gender, and arrows indicate the biplot vectors corresponding to the original features. Both methods reveal the underlying structure that separates male from female athletes, but in a different way. If for PCA all the features are roughly equally involved in the definition of the directions, only two features (body fat percentage \code{Bfat}, and weight \code{Wt}) are mainly associated with the estimated PPGMMGA directions.}
  \label{fig:ais}
\end{figure}

Figure~\ref{fig:ais} shows 2-dimensional projections obtained by applying the PPGMMGA method with the UT approximation and the PCA. Both graphs show two groups of data points, roughly corresponding to the athletes' gender, but with very different patterns. 
If for PCA all the variables are equally involved in the projection, for PPGMMGA two features appear to be the most important, i.e. \code{Bfat} and \code{Wt}. By looking at the left panel of Figure~\ref{fig:ais}, female athletes appear to have a higher body fat percentage compared to males. Furthermore, there are some outlying athletes (e.g. female gymnasts) with a relatively low weight.
Thus, even though the underlying structure of the data could not be represented by a Gaussian mixture, the proposed PPGMMGA procedure was able to largely unveil the clustering structure present within the data.

\subsection{Leukemia data}

This dataset originated from a study on gene expression from Affymetrix high-density oligonucleotide arrays \citep{Golub:etal:1999}. After preliminary screening and processing of genes, as described in \citet{Dudoit:Fridllyand:Speed:2002}, a matrix of gene expression levels for 3051 genes on 38 tumour mRNA samples was obtained. 27 samples were from class ALL (acute lymphoblastic leukemia) and 11 from class AML (acute myeloid leukemia). 
The leukemia dataset is an example of high-dimensional data having large $p$ and small $n$. \citet{Lee:Cook:2010} analysed the data using a supervised PP procedure with the aim of showing the different class structures based on an extension of the projection index proposed in \citet{Lee:Cook:Sigbert:Lumley:2005}.
The dataset is available in the \pkg{multtest} R package \citep{Pollard:Dudoit:vanDerLaan, multtest}. As preliminary step we only centred the data. 

Due to the large number of features (genes) compared to the relatively small number of observations (samples), only the diagonal covariances for the components were considered for density estimation using GMMs.
The best model according to BIC was the VVI model with 2 components.

\begin{figure}[!htb]
  \centering
  \includegraphics[width=0.9\textwidth]{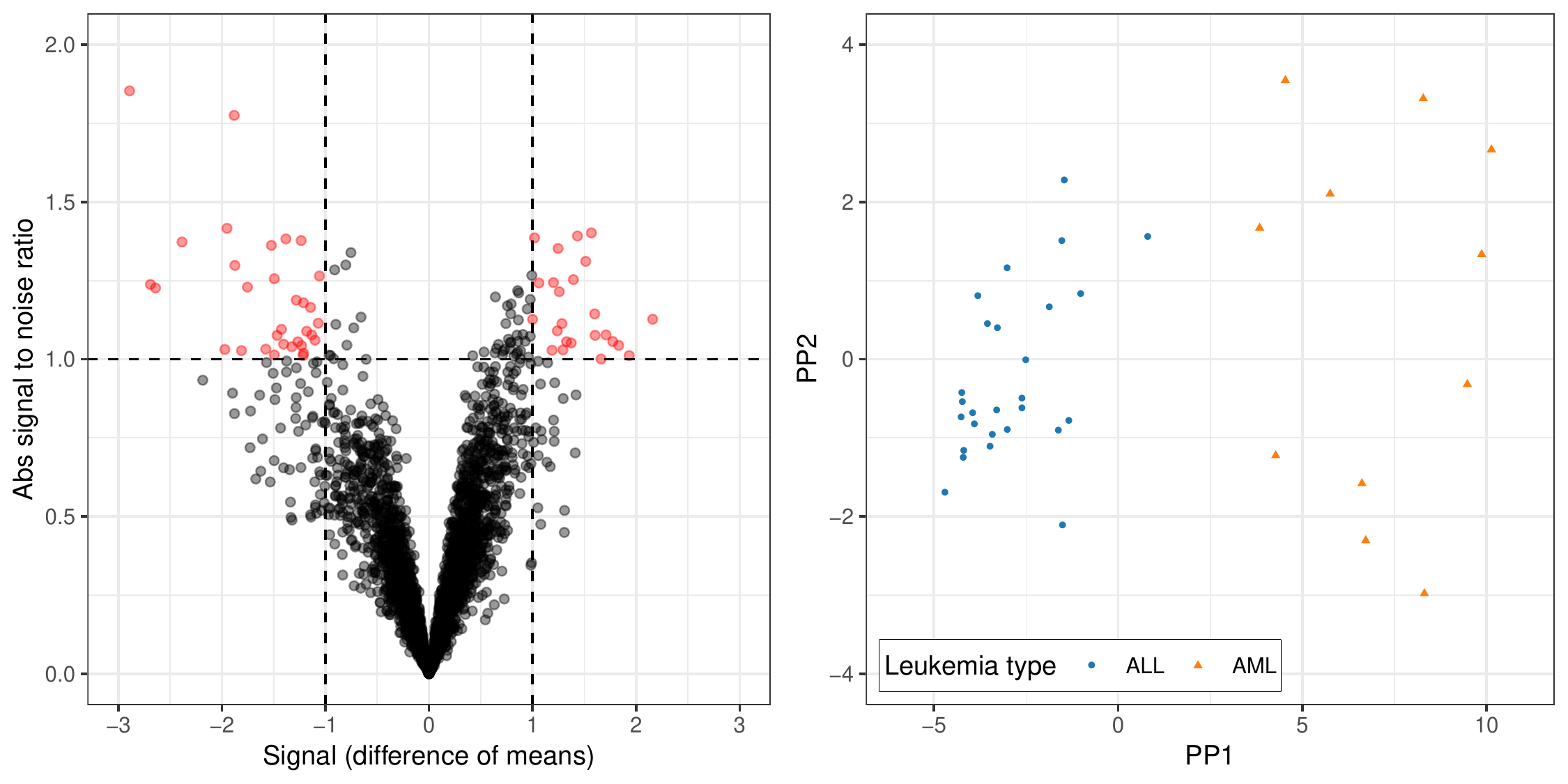}
  \caption{Volcano-type plot of signal vs absolute signal-to-noise ratio for selecting differentially expressed genes (left panel), and optimal 2-dimensional PPGMMGA projection (right panel). Points are marked according to tumour classes, acute lymphoblastic leukemia (ALL), and acute myeloid leukemia (AML). The two different type of tumours are clearly separated along the estimated PPGMMGA directions.}
  \label{fig:leukemia}
\end{figure}

A preliminary screening of the relevant genes was performed via a volcano-type plot \citep{Li:2012}. This was obtained by plotting the signal $\Signal_j = (\hat{\mu}_{j1} - \hat{\mu}_{j2})$, i.e. the difference of estimated component means $\hat{\mu}_{j1}$ and $\hat{\mu}_{j2}$ for each gene $j$, versus the absolute signal-to-noise ratio $\SNR_j = (\hat{\mu}_{j1} - \hat{\mu}_{j2})/(\hat{\sigma}_{j1} + \hat{\sigma}_{j2})$,
where $\hat{\sigma}_{j1}$ and $\hat{\sigma}_{j2}$ are the estimated component standard deviations of gene $j$ ($j=1,\ldots,3051$).
Differentially expressed genes should show both a signal and a signal-to-noise ratio relatively large (in absolute values).
By setting an arbitrary threshold, an informal gene selection procedure was conducted based on the volcano-type plot reported in the left panel of Figure~\ref{fig:leukemia}, which resulted in 56 selected genes.

Thus, a 2-dimensional PPGMMGA analysis based on the UT approximation was performed using the 56 selected genes, and produced the graph shown in the right panel of Figure~\ref{fig:leukemia}. From this graph we can see that the estimated PPGMMGA directions allow to easily separate the two type of leukemia tumours.

\section{Final comments}

In this work we have presented a new projection pursuit algorithm, called PPGMMGA, with the aim of visualising multivariate data when clustering structure is sought.

Preliminary visualisation is an important step to understand the data structure and may facilitate the specification and estimation of statistical models. On the other hand, projection pursuit procedures are computationally intensive and require the estimation of a projection density or a projection index which, in turn, needs maximisation algorithms to be able to achieve a global maximum.

The proposed methodology attempts to solve these limitations by exploiting some well know properties of GMMs, and using GAs for global optimisation.
Since we assume a flexible distributional form for the data, PPGMMGA can be considered a semi-parametric approach to projection pursuit.
Projection pursuit indices are introduced based on negentropy approximations and, due to higher accuracy and fast computation, we recommend the use of unscented transformation (UT) as the projection index.
Overall, the PPGMMGA procedure appears to work well on both simulated and real data, helping to detect the underlying clustering structure present in a dataset.

In future work, we would like to devote more attention to the selection of the dimension of the projection subspace. An incorrect choice of the dimensionality may lead to a sub-optimal visualisation of the data, unable to show any natural clusters. 
An analogous visualisation problem could also occur in the presence of a relatively large number of noise variables compared to the true "clustering features". In this case, in order to remove (at least a part of) the noise variables, a preliminary subset selection step could form the basis for an effective solution. Alternatively, a regularised PP index could be adopted. We defer a thorough study of these aspects to future research.

\section*{Supplemental Materials}

\begin{description}
\item[R code:] script file \code{code.R} containing R code for replicating the examples discussed in Section~\ref{sec:examples}.
\item[R package:] R package \code{ppgmmga} implementing the projection pursuit method described in the article.  The package is available on CRAN at \url{https://cran.r-project.org/package=ppgmmga}. 
\end{description}

\section*{Acknowledgements}
The authors are grateful to the Editor, the Associate Editor, and two anonymous Reviewers for their very helpful comments and suggestions which help to improve the paper.

\section*{Appendix}
\label{sec:appendix}

\paragraph{Proposition: linear transformation property of GMMs}\mbox{}\\
Let $\z = \b + \B\T\x$ be an affine transformation of $\x$, where $\B$ is a deterministic matrix, $\b$ is a deterministic vector, and $\x$ is a random vector distributed as a mixture of Gaussian distributions, i.e. $\x \sim  \sum^G_{g=1} \pi_g \phi(\x; \mub_g,\Sigmab_g)$. Then, the distribution of the linearly transformed random vector $\z$ is as follows:
\begin{equation*}
\z \sim \sum^{G}_{g=1} \pi_{g} \phi\left( \z ; \b + \B\T \mub_g, \B\T \Sigma_g \B \right),
\end{equation*}
\noindent\textit{Proof}.\\
Recall that the characteristic function of the Gaussian mixture random vector $\x$ is
\begin{equation*}
CF_{\x}(\t) = \sum_{g=1}^G \pi_g e^{i\t\T \mub_g - \frac{1}{2} \t\T \Sigmab_g \t}
\end{equation*}
for any real vector $\t$ and $i = \sqrt{-1}$.
Then, the characteristic function of the random vector $\z$ is as follows:
\begin{align*}
CF_{\z}(\t)
& = \E \left[ e^{i\t\T(\b + \B\T\x)} \right]
  = e^{i\t\T\b} \; \E \left[ e^{i(\B\t)\T\x} \right] \\
& = e^{i\t\T\b} \sum^{G}_{g=1} \pi_{g} e^{ i(\B\t)\T \mub_g - \frac{1}{2} (\B\t)\T \Sigmab_g (\B\t)\T } \\
& = \sum^{G}_{g=1} \pi_{g} e^{ i\t\T (\b + \B\T \mub_g) - \frac{1}{2} \t\T (\B\T \Sigmab_g \B)\t }.
\end{align*}
In the last equation we recognise the characteristic function of a Gaussian mixture distribution with component means $\b + \B\T \mub_g$ and component covariance matrices $\B\T \Sigmab_g \B$.

\bibliographystyle{apalike}
\bibliography{ppgmmga_arxiv}

\end{document}